\documentclass[conference]{IEEEtran}
\IEEEoverridecommandlockouts
\usepackage{amsmath,amssymb,amsfonts}
\usepackage{algorithmic}
\usepackage{graphicx}
\usepackage{textcomp}
\usepackage{array}
\usepackage{xcolor}
\usepackage{booktabs}
\usepackage[normalem]{ulem}
\def\BibTeX{{\rm B\kern-.05em{\sc i\kern-.025em b}\kern-.08em
    T\kern-.1667em\lower.7ex\hbox{E}\kern-.125emX}}
\usepackage{authblk} 

\usepackage[backend=biber, sorting=none, style=numeric-comp, maxnames=5]{biblatex}
\usepackage[switch]{lineno}
\usepackage{acronym}

\DeclareSourcemap{
  \maps[datatype=bibtex]{
    \map[overwrite=true]{
      \step[fieldset=url, null]
      \step[fieldset=eprint, null]
    }
  }
}

\acrodef{ADC}[ADC]{Analog-to-Digital Converter}
\acrodef{ADEXP}[AdExp-IF]{Adaptive Exponential Integrate-and-Fire}
\acrodef{ADM}[ADM]{Asynchronous Delta Modulator}
\acrodef{AER}[AER]{Address-Event Representation}
\acrodef{AEX}[AEX]{AER EXtension board}
\acrodef{AE}[AE]{Address-Event}
\acrodef{AFE}[AFE]{Analog Front-End}
\acrodef{AFM}[AFM]{Atomic Force Microscope}
\acrodef{AGC}[AGC]{Automatic Gain Control}
\acrodef{AI}[AI]{Artificial Intelligence}
\acrodef{AMDA}[AMDA]{AER Motherboard with D/A converters}
\acrodef{AMPA}[AMPA]{$\alpha$-Amino-3-hydroxy-5-methyl-4-isoxazolepropionic Acid}
\acrodef{ANN}[ANN]{Artificial Neural Network}
\acrodef{API}[API]{Application Programming Interface}
\acrodef{APMOM}[APMOM]{Alternate Polarity Metal On Metal}
\acrodef{ARM}[ARM]{Advanced RISC Machine}
\acrodef{ASIC}[ASIC]{Application Specific Integrated Circuit}
\acrodef{BCM}[BMC]{Bienenstock-Cooper-Munro}
\acrodef{BD}[BD]{Bundled Data}
\acrodef{BEOL}[BEOL]{Back-end of Line}
\acrodef{BG}[BG]{Bias Generator}
\acrodef{BMI}[BMI]{Brain-Machince Interface}
\acrodef{BTB}[BTB]{Band-to-Band tunnelling}
\acrodef{BTSP}[BTSP]{Behavioral Time Scale Synaptic Plasticity}
\acrodef{CAD}[CAD]{Computer Aided Design}
\acrodef{CAM}[CAM]{Content Addressable Memory}
\acrodef{CAVIAR}[CAVIAR]{Convolution AER Vision Architecture for Real-Time}
\acrodef{CA}[CA]{Cortical Automaton}
\acrodef{CCN}[CCN]{Cooperative and Competitive Network}
\acrodef{CDR}[CDR]{Clock-Data Recovery}
\acrodef{CFC}[CFC]{Current to Frequency Converter}
\acrodef{CHP}[CHP]{Communicating Hardware Processes}
\acrodef{CMIM}[CMIM]{Metal-Insulator-Metal Capacitor}
\acrodef{CML}[CML]{Current Mode Logic}
\acrodef{CMOL}[CMOL]{Hybrid CMOS nanoelectronic circuits}
\acrodef{CMOS}[CMOS]{Complementary Metal-Oxide-Semiconductor}
\acrodef{CNN}[CNN]{Convolutional Neural Network}
\acrodef{CNS}[CNS]{central Nervous System}
\acrodef{COTS}[COTS]{Commercial Off-The-Shelf}
\acrodef{CPG}[CPG]{Central Pattern Generator}
\acrodef{CPLD}[CPLD]{Complex Programmable Logic Device}
\acrodef{CPU}[CPU]{Central Processing Unit}
\acrodef{CSM}[CSM]{Cortical State Machine}
\acrodef{CSP}[CSP]{Constraint Satisfaction Problem}
\acrodef{CTXCTL}[CTXCTL]{CortexControl}
\acrodef{CV}[CV]{Coefficient of Variation}
\acrodef{DAC}[DAC]{Digital to Analog Converter}
\acrodef{DAS}[DAS]{Dynamic Auditory Sensor}
\acrodef{DAVIS}[DAVIS]{Dynamic and Active Pixel Vision Sensor}
\acrodef{DBN}[DBN]{Deep Belief Network}
\acrodef{DBS}[DBS]{Deep Brain Stimulation}
\acrodef{DFA}[DFA]{Deterministic Finite Automaton}
\acrodef{DIBL}[DIBL]{Drain-Induced Barrier-Lowering}
\acrodef{DI}[DI]{Delay Insensitive}
\acrodef{DMA}[DMA]{Direct Memory Access}
\acrodef{DNF}[DNF]{Dynamic Neural Field}
\acrodef{DNN}[DNN]{Deep Neural Network}
\acrodef{DOF}[DOF]{Degrees of Freedom}
\acrodef{DPE}[DPE]{Dynamic Parameter Estimation}
\acrodef{DPI}[DPI]{Differential Pair Integrator}
\acrodef{DRAM}[DRAM]{Dynamic Random Access Memory}
\acrodef{DRRZ}[DR-RZ]{Dual-Rail Return-to-Zero}
\acrodef{DR}[DR]{Dual Rail}
\acrodef{DSP}[DSP]{Digital Signal Processor}
\acrodef{DVS}[DVS]{Dynamic Vision Sensor}
\acrodef{DYNAP}[DYNAP]{Dynamic Neuromorphic Asynchronous Processor}
\acrodef{EBL}[EBL]{Electron Beam Lithography}
\acrodef{ECG}[ECG]{Electrocardiography}
\acrodef{ECoG}[ECoG]{Electrocorticography}
\acrodef{EDVAC}[EDVAC]{Electronic Discrete Variable Automatic Computer}
\acrodef{EEG}[EEG]{Electroencephalography}
\acrodef{EI}[EI]{Excitatory-Inhibitory}
\acrodef{EIN}[EIN]{Excitatory-Inhibitory Network}
\acrodef{EMG}[EMG]{Electromyography}
\acrodef{EM}[EM]{Expectation Maximization}
\acrodef{EOG}[EOG]{Electrooculogram}
\acrodef{EPSC}[EPSC]{Excitatory Post-Synaptic Current}
\acrodef{EPSP}[EPSP]{Excitatory Post-Synaptic Potential}
\acrodef{EZ}[EZ]{Epileptogenic Zone}
\acrodef{FDSOI}[FDSOI]{Fully-Depleted Silicon on Insulator}
\acrodef{FET}[FET]{Field-Effect Transistor}
\acrodef{FFT}[FFT]{Fast Fourier Transform}
\acrodef{FI}[F-I]{Frequency--Current}
\acrodef{FMA}[FMA]{Floating Microelectrode Array}
\acrodef{FNN}[FNN]{Feed-forward Neural Network}
\acrodef{FPGA}[FPGA]{Field Programmable Gate Array}
\acrodef{FR}[FR]{Fast Ripple}
\acrodef{FSA}[FSA]{Finite State Automaton}
\acrodef{FSM}[FSM]{Finite State Machine}
\acrodef{GABA}[GABA]{$\gamma$-Aminobutanoic Acid}
\acrodef{GIDL}[GIDL]{Gate-Induced Drain Leakage}
\acrodef{GOPS}[GOPS]{Giga-Operations per Second}
\acrodef{GPIO}[GPIO]{General Purpose I/O}
\acrodef{GPU}[GPU]{Graphical Processing Unit}
\acrodef{GT}[GT]{Ground Truth}
\acrodef{GUI}[GUI]{Graphical User Interface}
\acrodef{HAL}[HAL]{Hardware Abstraction Layer}
\acrodef{HFO}[HFO]{High Frequency Oscillation}
\acrodef{HH}[H\&H]{Hodgkin \& Huxley}
\acrodef{HMM}[HMM]{Hidden Markov Model}
\acrodef{HRS}[HRS]{High-Resistive State}
\acrodef{HR}[HR]{Heart Rate}
\acrodef{HSE}[HSE]{Handshaking Expansion}
\acrodef{HW}[HW]{Hardware}
\acrodef{ICT}[ICT]{Information and Communication Technology}
\acrodef{IC}[IC]{Integrated Circuit}
\acrodef{IF2DWTA}[IF2DWTA]{Integrate \& Fire 2-Dimensional WTA}
\acrodef{IFSLWTA}[IFSLWTA]{Integrate \& Fire Stop Learning WTA}
\acrodef{IF}[I\&F]{Integrate-and-Fire}
\acrodef{IMU}[IMU]{Inertial Measurement Unit}
\acrodef{INCF}[INCF]{International Neuroinformatics Coordinating Facility}
\acrodef{INI}[INI]{Institute of Neuroinformatics}
\acrodef{IO}[I/O]{Input/Output}
\acrodef{IPSC}[IPSC]{Inhibitory Post-Synaptic Current}
\acrodef{IPSP}[IPSP]{Inhibitory Post-Synaptic Potential}
\acrodef{IP}[IP]{Intellectual Property}
\acrodef{ISI}[ISI]{Inter-Spike Interval}
\acrodef{IoT}[IoT]{Internet of Things}
\acrodef{JFLAP}[JFLAP]{Java - Formal Languages and Automata Package}
\acrodef{LEDR}[LEDR]{Level-Encoded Dual-Rail}
\acrodef{LFP}[LFP]{Local Field Potential}
\acrodef{LGBM}[LightGBM]{Light Gradient Boosting Machine}
\acrodef{LIFE}[LIFE]{Longitudinal Intrafascicular Electrodes}
\acrodef{LIF}[LI\&F]{Leaky Integrate-and-Fire}
\acrodef{LLC}[LLC]{Low Leakage Cell}
\acrodef{LNA}[LNA]{Low-Noise Amplifier}
\acrodef{LPF}[LPF]{Low Pass Filter}
\acrodef{LRS}[LRS]{Low-Resistive State}
\acrodef{LSM}[LSM]{Liquid State Machine}
\acrodef{LTD}[LTD]{Long Term Depression}
\acrodef{LTI}[LTI]{Linear Time-Invariant}
\acrodef{LTP}[LTP]{Long Term Potentiation}
\acrodef{LTU}[LTU]{Linear Threshold Unit}
\acrodef{LUT}[LUT]{Look-Up Table}
\acrodef{LVDS}[LVDS]{Low Voltage Differential Signaling}
\acrodef{MCMC}[MCMC]{Markov-Chain Monte Carlo}
\acrodef{MEA}[MEA]{Multielectrode Arrays}
\acrodef{MEMS}[MEMS]{Micro Electro Mechanical System}
\acrodef{MFR}[MFR]{Mean Firing Rate}
\acrodef{MIM}[MIM]{Metal Insulator Metal}
\acrodef{MLP}[MLP]{Multilayer Perceptron}
\acrodef{ML}[ML]{Machine Learning}
\acrodef{MOSCAP}[MOSCAP]{Metal Oxide Semiconductor Capacitor}
\acrodef{MOSFET}[MOSFET]{Metal Oxide Semiconductor Field-Effect Transistor}
\acrodef{MOS}[MOS]{Metal Oxide Semiconductor}
\acrodef{MRI}[MRI]{Magnetic Resonance Imaging}
\acrodef{NCS}[NCS]{Neuromorphic Cognitive Systems}
\acrodef{NDFSM}[NDFSM]{Non-deterministic Finite State Machine}
\acrodef{ND}[ND]{Noise-Driven}
\acrodef{NEF}[NEF]{Neural Engineering Framework}
\acrodef{NHML}[NHML]{Neuromorphic Hardware Mark-up Language}
\acrodef{NIL}[NIL]{Nano-Imprint Lithography}
\acrodef{NI}[NI]{Neural Interface}
\acrodef{NMDA}[NMDA]{\textit{N}-Methyl-\textsc{d}-aspartate}
\acrodef{NME}[NE]{Neuromorphic Engineering}
\acrodef{NN}[NN]{Neural Network}
\acrodef{NOC}[NoC]{Network-on-Chip}
\acrodef{NRZ}[NRZ]{Non-Return-to-Zero}
\acrodef{NSM}[NSM]{Neural State Machine}
\acrodef{OR}[OR]{Operating Room}
\acrodef{OTA}[OTA]{Operational Transconductance Amplifier}
\acrodef{PCB}[PCB]{Printed Circuit Board}
\acrodef{PCHB}[PCHB]{Pre-Charge Half-Buffer}
\acrodef{PCM}[PCM]{Phase Change Memory}
\acrodef{PC}[PC]{Personal Computer}
\acrodef{PDK}[PDK]{Process Design Kit}
\acrodef{PE}[PE]{Phase Encoding}
\acrodef{PFA}[PFA]{Probabilistic Finite Automaton}
\acrodef{PFC}[PFC]{Prefrontal Cortex}
\acrodef{PFM}[PFM]{Pulse Frequency Modulation}
\acrodef{PNI}[PNI]{Peripheral Nerve Interface}
\acrodef{PNS}[PNS]{Peripheral Nervous System}
\acrodef{PPG}[PPG]{Photoplethysmography}
\acrodef{PR}[PR]{Production Rule}
\acrodef{PSC}[PSC]{Post-Synaptic Current}
\acrodef{PSP}[PSP]{Post-Synaptic Potential}
\acrodef{PSTH}[PSTH]{Peri-Stimulus Time Histogram}
\acrodef{PV}[PV]{Parvalbumin}
\acrodef{PYR}[PYR]{Pyramidal}
\acrodef{QDI}[QDI]{Quasi Delay Insensitive}
\acrodef{RAM}[RAM]{Random Access Memory}
\acrodef{RA}[RA]{Resected Area}
\acrodef{RDF}[RDF]{Random Dopant Fluctuation}
\acrodef{RELU}[ReLu]{Rectified Linear Unit}
\acrodef{RISC}[RISC]{Reduced Instruction Set Computer}
\acrodef{RLS}[RLS]{Recursive Least-Squares}
\acrodef{RMSE}[RMSE]{Root Mean Square-Error}
\acrodef{RMS}[RMS]{Root Mean Square}
\acrodef{RNN}[RNN]{Recurrent Neural Network}
\acrodef{ROLLS}[ROLLS]{Reconfigurable On-Line Learning Spiking}
\acrodef{RRAM}[R-RAM]{Resistive Random Access Memory}
\acrodef{RSA}[RSA]{Respiratory Sinus Arrhythmia}
\acrodef{R}[R]{Ripple}
\acrodef{SAC}[SAC]{Selective Attention Chip}
\acrodef{SAT}[SAT]{Boolean Satisfiability Problem}
\acrodef{SCI}[SCI]{Spinal Cord Injury}
\acrodef{SCX}[SCX]{Silicon CorteX}
\acrodef{SD}[SD]{Signal-Driven}
\acrodef{SEM}[SEM]{Spike-based Expectation Maximization}
\acrodef{SLAM}[SLAM]{Simultaneous Localization and Mapping}
\acrodef{SMP}[SMP]{Soil Metric Potential}
\acrodef{SNN}[SNN]{Spiking Neural Network}
\acrodef{SNR}[SNR]{Signal to Noise Ratio}
\acrodef{SOC}[SoC]{System-On-Chip}
\acrodef{SOI}[SOI]{Silicon on Insulator}
\acrodef{SOZ}[SOZ]{Seizure Onset Zone}
\acrodef{SPI}[SPI]{Serial Peripheral Interface}
\acrodef{SP}[SP]{Separation Property}
\acrodef{SRAM}[SRAM]{Static Random Access Memory}
\acrodef{SST}[SST]{Somatostatin}
\acrodef{STDP}[STDP]{Spike-Timing Dependent Plasticity}
\acrodef{STD}[STD]{Short-Term Depression}
\acrodef{STP}[STP]{Short-Term Plasticity}
\acrodef{STT-MRAM}[STT-MRAM]{Spin-Transfer Torque Magnetic Random Access Memory}
\acrodef{STT}[STT]{Spin-Transfer Torque}
\acrodef{SVM}[SVM]{Support Vector Machine}
\acrodef{SW}[SW]{Software}
\acrodef{TCAM}[TCAM]{Ternary Content-Addressable Memory}
\acrodef{TFT}[TFT]{Thin Film Transistor}
\acrodef{TIME}[TIME]{Transverse Intrafascicular Multichannel Electrode}
\acrodef{TLE}[TLE]{Temporal Lobe Epilepsy}
\acrodef{UEA}[UEA]{Utah Electrode Array}
\acrodef{USB}[USB]{Universal Serial Bus}
\acrodef{USEA}[USEA]{Utah Slanted Electrode Array}
\acrodef{VHDL}[VHDL]{VHSIC Hardware Description Language}
\acrodef{VHSIC}[VHSIC]{Very High Speed Integrated Circuits}
\acrodef{VIP}[VIP]{Vasoactive Intestinal Peptide}
\acrodef{VLSI}[VLSI]{Very Large Scale Integration}
\acrodef{VNS}[VNS]{Vagal Nerve Stimulation}
\acrodef{VOR}[VOR]{Vestibulo-Ocular Reflex}
\acrodef{VSA}[VSA]{Vector Symbolic Architecture}
\acrodef{WCST}[WCST]{Wisconsin Card Sorting Test}
\acrodef{WTA}[WTA]{Winner-Take-All}
\acrodef{XML}[XML]{eXtensible Mark-up Language}
\acrodef{divmod3}[DIVMOD3]{Divisibility of a number by three}
\acrodef{hWTA}[hWTA]{Hard Winner-Take-All}
\acrodef{iEEG}[iEEG]{Intracranial Electroencephalography}
\acrodef{rSNN}[rSNN]{recurrent Spiking Neural Network}
\acrodef{sWTA}[sWTA]{soft Winner-Take-All}

\title{Queen Detection in Beehives via Environmental Sensor Fusion for Low-Power Edge Computing}

\author[1,2,$\dagger$]{Chiara De Luca}
\author[1,2]{Elisa Donati}

\affil[1]{Institute of Neuroinformatics\\University of Zurich and ETH Zurich, Zurich, Switzerland}
\affil[2]{Digital Society Initiative\\University of Zurich, Zurich, Switzerland}
\affil[$\dagger$]{corresponding author: \emph{chiaradeluca@ini.uzh.ch}}

\begin{document}
\maketitle
\IEEEpubidadjcol 
\begin{abstract}
Queen bee presence is essential for the health and stability of honeybee colonies, yet current monitoring methods rely on manual inspections that are labor-intensive, disruptive, and impractical for large-scale beekeeping. While recent audio-based approaches have shown promise, they often require high power consumption, complex preprocessing, and are susceptible to ambient noise. To overcome these limitations, we propose a lightweight, multimodal system for queen detection based on environmental sensor fusion—specifically, temperature, humidity, and pressure differentials between the inside and outside of the hive. Our approach employs quantized decision tree inference on a commercial STM32 microcontroller, enabling real-time, low-power edge computing without compromising accuracy. We show that our system achieves over 99\% queen detection accuracy using only environmental inputs, with audio features offering no significant performance gain. This work presents a scalable and sustainable solution for non-invasive hive monitoring, paving the way for autonomous, precision beekeeping using off-the-shelf, energy-efficient hardware.
\end{abstract}


\section{Introduction}
\label{sec:intro}
Honeybee colonies are critical for biodiversity and food production, but are increasingly threatened by environmental stressors~\cite{Potts_etal16}. Monitoring the presence of the queen bee is particularly critical, as her absence can rapidly destabilize the colony and lead to collapse~\cite{Barron_etal15}. Manual inspections are laborious and disruptive, motivating the need for automated, non-invasive detection methods.

Recent advancements in beehive monitoring have increasingly embraced multi-sensor approaches to detect critical colony events such as queen loss. One notable system combines continuous measurements of temperature, humidity, and acoustic signals to infer queen absence based on internal hive conditions and behavioral cues~\cite{Lu_etal24}. Extending this direction, the Multi-modal Sensor dataset with Phenotypic trait measurements from honey Bees (MSPB) provides a rich, longitudinal resource of synchronized audio and environmental data from 53 hives. Expert annotations—including queen cell presence—enable advanced machine learning analyses of colony status~\cite{Zhu_eta24}. Complementing these data-driven initiatives, the BHiveSense platform introduces a modular, microservice-based IoT architecture that integrates environmental, structural, and acoustic data—such as temperature, humidity, weight, lid status, and sound—for real-time decision-making and sustainable apiary management~\cite{Cota_etal23}. Similarly, a real-world deployment demonstrated a multi-sensor system capturing acoustic, microclimatic (temperature, humidity, $CO_2$), weight, and weather data to monitor hive health and forecast anomalies—highlighting the value of comprehensive sensing for early detection of colony disruptions~\cite{borowik2019multisensor}.

At the same time, audio-only approaches have gained traction due to the queen’s distinctive acoustic signatures. While effective in controlled settings, these methods typically require complex preprocessing pipelines and are sensitive to ambient noise and microphone placement~\cite{De_etal24}. Despite efforts to port such systems to microcontrollers using TinyML frameworks, continuous audio acquisition remains energy-intensive due to the demands of recording, buffering, and spectral feature extraction. Moreover, acoustic signals can be unreliable during passive hive phases or in noisy environments. Even TinyML-based implementations are constrained by the need for audio chunking and buffering, limiting their scalability for real-time, embedded applications.

Our work bridges these domains by proposing a lightweight, audio-optional architecture that leverages differential environmental sensing—specifically, temperature, humidity, and pressure variations between the hive interior and exterior—as a robust, low-power indicator of queen presence. Unlike prior systems that depend on audio or perform sensor fusion only offline, our solution supports real-time embedded inference on microcontrollers. This enables scalable, interpretable, and energy-efficient monitoring suitable for long-term field deployment.

The key innovation lies in demonstrating that high queen detection accuracy ($>$99\%) can be achieved using only low-cost environmental sensors, with audio playing at most a marginal role. This marks a shift from sound-intensive methods to explainable, low-power sensor fusion architectures. To validate our approach, we implement a two-part system: a Python-based training pipeline and a real-time embedded deployment on an STM32 microcontroller. The model—converted into quantized decision tree logic—achieves high accuracy with minimal memory and computational overhead.

This work lays the foundation for scalable, sustainable hive monitoring systems that are practical, non-invasive, and compatible with edge devices—opening towards autonomous precision beekeeping and embedded ecological sensing.

\section{Methods}
\label{sec:methods}
\begin{figure}
\centering
\includegraphics[width=0.9\linewidth]{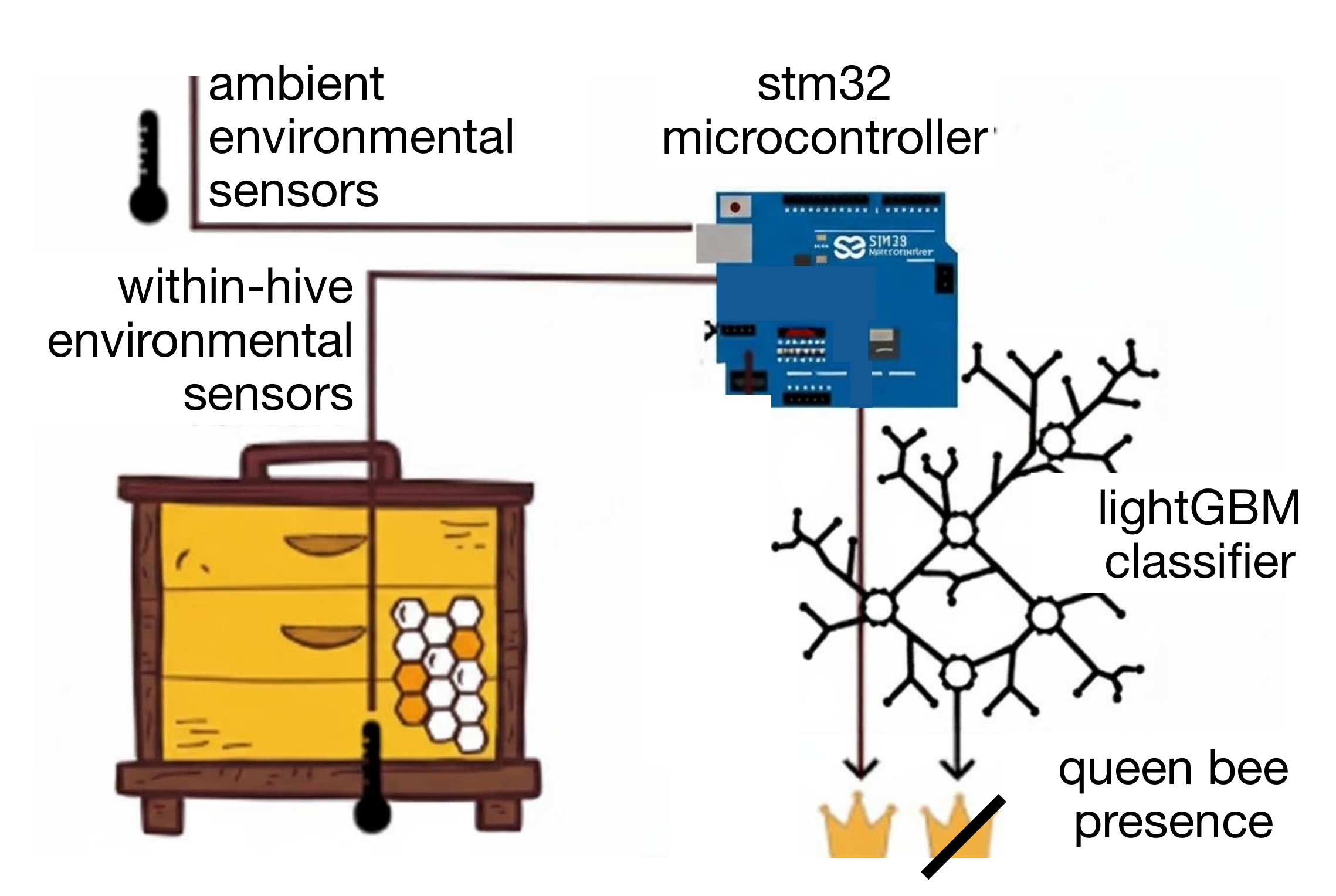}
\caption{System overview: environmental data are acquired inside and outside the hive, processed on the edge by a LightGBM system on an STM32 to classify the precence of the queen bee within the hive.}
\label{fig:overview}
\end{figure}

The proposed system is composed of two primary components: a Python-based data processing pipeline and a real-time prediction system implemented on an STM32 microcontroller. The Python component is responsible for tasks such as data preprocessing, model training, and performance validation. In parallel, the STM32 system handles real-time operations, including environmental data acquisition, feature computation, and predictive inference (Fig.~\ref{fig:overview}), enabling offline training combined with low-latency, embedded real-time deployment.

\subsection{Dataset}
This study used the Bee Audio Dataset~\cite{anna_yang_2022}, a publicly available resource comprising recordings from multiple beehives. Together with audio signals, each data point includes timestamped environmental measurements—temperature, humidity, and pressure—collected both inside and outside the hive.

In addition to the environmental sensor data, each entry is annotated with metadata detailing hive conditions and behavioral observations. Of particular interest to this study is the presence or absence of the queen bee, which serves as the ground truth label for training and validating the classification model. This binary label is critical for understanding colony dynamics and is the primary target for prediction. It is important to note that the dataset comprises recordings from only four hives, with a significantly unbalanced distribution of queen presence vs. absence events. As a result, the same hive may be represented in both the training and test sets, effectively simulating a scenario where the model benefits from a form of 'pre-training' on that specific hive prior to deployment. While this setup is useful for proof-of-concept and in-hive applications, future work should focus on expanding the dataset to support generalization to unseen hives, which is essential for truly scalable and robust hive monitoring systems.

\subsection{Environmental Data Processing}
Only environmental sensor measurements were used for model input. These included temperature, humidity, and pressure values, each recorded simultaneously inside and outside the hive. Rather than using raw sensor readings, we computed differential features to capture relative changes between internal hive conditions and the ambient environment. 
These differentials are hypothesized to be more indicative of hive status and internal activity than absolute values, as they reflect the hive's regulation capabilities in response to external conditions.

\subsection{Machine Learning Model}
To detect the presence of the queen bee, we employed a \ac{LGBM} classifier~\cite{fan2019light} on environmental differential features, with attention paid to class balance and overfitting prevention through appropriate sampling and validation techniques.
The complete dataset was divided into training and test subsets using an 80/20 stratified split to preserve the original class balance. Also, we applied 5-fold cross-validation on the training set. This approach allowed us to fine-tune hyperparameters and obtain an unbiased estimate of the model’s generalization performance. Stratification during both splitting and cross-validation guaranteed that minority classes were adequately represented throughout the training and evaluation pipeline.
The  \ac{LGBM} model was trained using standard hyperparameter tuning via 5-fold cross-validation, with early stopping to prevent overfitting. Built-in class weighting addressed data imbalance, and default parameters were used unless otherwise noted.
Training was conducted using binary cross-entropy as the objective function, with early stopping employed to halt training after 10 rounds of no improvement in validation loss. All experiments have been repeated 10 times with different random-seed initializations.
The best-performing model from the cross-validation phase was selected for deployment on the STM32 microcontroller. To make the model compatible with the embedded environment, we converted the trained decision trees into a C-compatible format. Feature scaling parameters were also extracted and implemented on the microcontroller to match the preprocessing performed during training.

\subsection{Real-time Implementation}
The STM32 microcontroller~\cite{stm32} was programmed for real-time data acquisition, feature extraction from environmental inputs, and model inference, outputting a binary classification of queen presence. The embedded software uses UART to receive sensor readings, construct feature vectors from differential values, and run inference. 
%
To facilitate efficient communication between the host computer and the STM32 microcontroller, a custom data transmission protocol was implemented. Environmental sensor readings were transmitted as 32-bit floating-point values. A designated marker was used to indicate the end of each data transmission sequence, ensuring data synchronization and completeness.

\subsection{STM32 Power Consumption Measurement}
\label{ssec:power}
To assess the energy efficiency of our embedded system, we evaluated its performance on a commercial STM32 NUCLEO-F767ZI microcontroller operating at 48 MHz and 3.3 V. Based on standard STM32 power profiles and our recorded processing time of $\sim$2.2 seconds per inference, the average power consumption is estimated at $\sim$44.5 mW, resulting in an energy cost of approximately 98 mJ per inference. This represents a significant improvement in efficiency compared to audio-based TinyML systems, which typically exceed 600 mJ per inference. Despite the reduced clock speed, our system maintains $>$99\% accuracy and achieves real-time capability for practical monitoring intervals. These results confirm the suitability of our lightweight, sensor-fusion-based approach for low-power, long-term deployment in field conditions.

\section{Results}
\label{sec:results}
\begin{table}[!h]
\centering
\caption{Performance using different environmental feature combinations.}
\label{tab:env-performance}
\begin{tabular}{lcc}
\toprule
\textbf{Feature Set} & \textbf{Val. Accuracy (\%)} & \textbf{Test Accuracy (\%)}\\
\midrule
Temperature & $86.8 \pm 0.2$ & $86.9 \pm 0.5$\\
Humidity & $86.8 \pm 0.1$ & $86.8 \pm 0.5$ \\
Pressure & $86.5 \pm 0.2$ & $86.7 \pm 0.3$\\
Temp + Humidity & $96.7 \pm 0.3$ & $97.2 \pm 0.5$\\
Temp + Pressure & $96.2 \pm 0.4$ & $96.8 \pm 0.7$\\
Humidity + Pressure & $95.5 \pm 0.4$ & $95.8 \pm 0.6$\\
All-Environmental & $\mathbf{99.0 \pm 0.2}$ & $\mathbf{99.4 \pm 0.4}$\\
Environmental + audio & $98.0 \pm 0.3$ & $98.7 \pm 0.6$\\
\bottomrule
\end{tabular}
\end{table}


\subsection{Classification perfomances on environmental features}
\label{ssec:classification}
The proposed queen detection system showed strong performance using only environmental features derived from the difference between internal and external hive conditions. Models trained on individual environmental features---specifically temperature, humidity, or pressure differentials---achieved average test accuracies of approximately 86.7\% to 86.9\%. While each sensor provides useful information on its own, their standalone predictive power is limited.
Performance improved substantially when environmental features were combined. The model trained with both temperature and humidity differentials achieved a test accuracy of 97.2\%, followed by temperature and pressure (96.8\%) and humidity and pressure (95.8\%). These results highlight the importance of combining complementary measurements to improve classification accuracy.

The best performance was obtained using all three environmental features. This model reached an average validation accuracy of 99.0\% and an average test accuracy of 99.4\%, with relatively balanced feature importance values across the three differentials. This confirms that multi-sensor integration  enhances the system's ability to detect queen presence.

Feature importance values obtained from the \ac{LGBM} model revealed that all three environmental differentials contributed meaningfully to the prediction. In the full model, humidity differential had the highest importance, followed by temperature and pressure (Table~\ref{tab:env-performance}).
\begin{table}[h!]
\centering
\caption{Classification report: python inference vs STM32 inference}
\begin{minipage}{0.48\textwidth}
\centering
\textbf{Python \text{\textbar} STM32 Inference}

\begin{tabular}{lcccc}
\toprule
Class & Precision & Recall & F1-score & Support \\
\midrule
0 & 1.00 \text{\textbar} 1.00 & 0.96 \text{\textbar} 0.94 & 0.98 \text{\textbar} 0.97 & 188\\
1 & 0.99 \text{\textbar} 0.99 & 1.00 \text{\textbar} 1.00 & 1.00 \text{\textbar} 1.00 & 1232\\
\midrule
Accuracy  & \multicolumn{4}{c}{0.99 \text{\textbar} 0.99}  \\
Macro avg & 1.00 \text{\textbar} 1.00 & 0.98 \text{\textbar} 0.97 & 0.99 \text{\textbar} 0.98 & 1420 \\
Weighted avg & 0.99 \text{\textbar} 0.99 & 0.99 \text{\textbar} 0.99 & 0.99 \text{\textbar} 0.99 & 1420\\
\bottomrule
\end{tabular}
\end{minipage}
\label{tab:stm-performance}
\end{table}

\begin{table*}[h!]
\centering
\caption{Comparison of Beehive Monitoring Studies}
\begin{tabular}{
    >{\raggedright\arraybackslash}p{1.1cm} 
    | >{\raggedright\arraybackslash}p{1.7cm} 
    | >{\raggedright\arraybackslash}p{1.9cm} 
    | >{\raggedright\arraybackslash}p{1.6cm} 
    | >{\raggedright\arraybackslash}p{1.7cm} 
    | >{\raggedright\arraybackslash}p{2cm}
    | >{\raggedright\arraybackslash}p{2.3cm}
    | >{\raggedright\arraybackslash}p{2.6cm}
}
\toprule
\textbf{Parameter} & \textbf{De Simone et al. (2024)~\cite{De_etal24}} & \textbf{Our Work} & \textbf{Bricout et al (2024)~\cite{Bricout_etal24}} & \textbf{Abdollahi et al (2025)~\cite{Abdollahi_etal25}} & \textbf{Sakova et al. (2024)~\cite{Sakova_etal24}} & \textbf{Barbisan et al. (2023)~\cite{Barbisan_etal23}} & \textbf{De Simone et al. (2024b)~\cite{DeSimone_etal24}} \\
\midrule
\midrule
\textbf{Sensor Type} &
Audio, temp, humidity &
Temp, humidity, pressure &
Audio, temp, humidity &
Audio,~SHTC1 temp/humidity &
Audio, temp, humidity &
Audio &
Audio \\
\midrule
\textbf{Approach} &
TinyML &
LightGBM &
CNN &
RandomForest &
1DCNN/LSTM &
SVM and NN &
2 layers-NN  \\
\midrule
\textbf{Accuracy} &
91–95\%  &
$>$99\% &
$\sim$90\% &
96.49\%  &
96.49\% (CNN), 84.28\% (LSTM) &
Up to 99.6\% (NN), 98.0\% (SVM) &
Up to 96.6\% (MFCC), 96.5\% (STFT) \\
\midrule
\textbf{Platform} &
ArduinoNano &
STM32 &
Raspberry Pi&
LocalServers &
MATLAB &
CPU &
STM32, TinyML \\
\midrule
\textbf{Power per Inf.} &
Not reported &
max. 92.4 mW &
Not reported &
Not reported &
Not reported &
Not reported &
STFT $\sim$11–12 mA @ 3.6V, MFCC $\sim$12–13 mA @ 3.6V\\
\midrule
\textbf{Energy per Inf.} &
$\sim$400~mJ &
max 6.45 mJ &
Not reported &
Not reported &
Not reported &
Not reported &
STFT (3s) 74.4 mJ, MFCC 70.8 mJ \\
\bottomrule
\end{tabular}
\label{tab:comparison}
\end{table*}

These results emphasize the benefits of sensor fusion: using a combination of environmental features provides a more complete representation of hive conditions, resulting in higher predictive accuracy and model robustness.
\begin{figure}[!h]
    \centering
    \includegraphics[width=1.\linewidth]{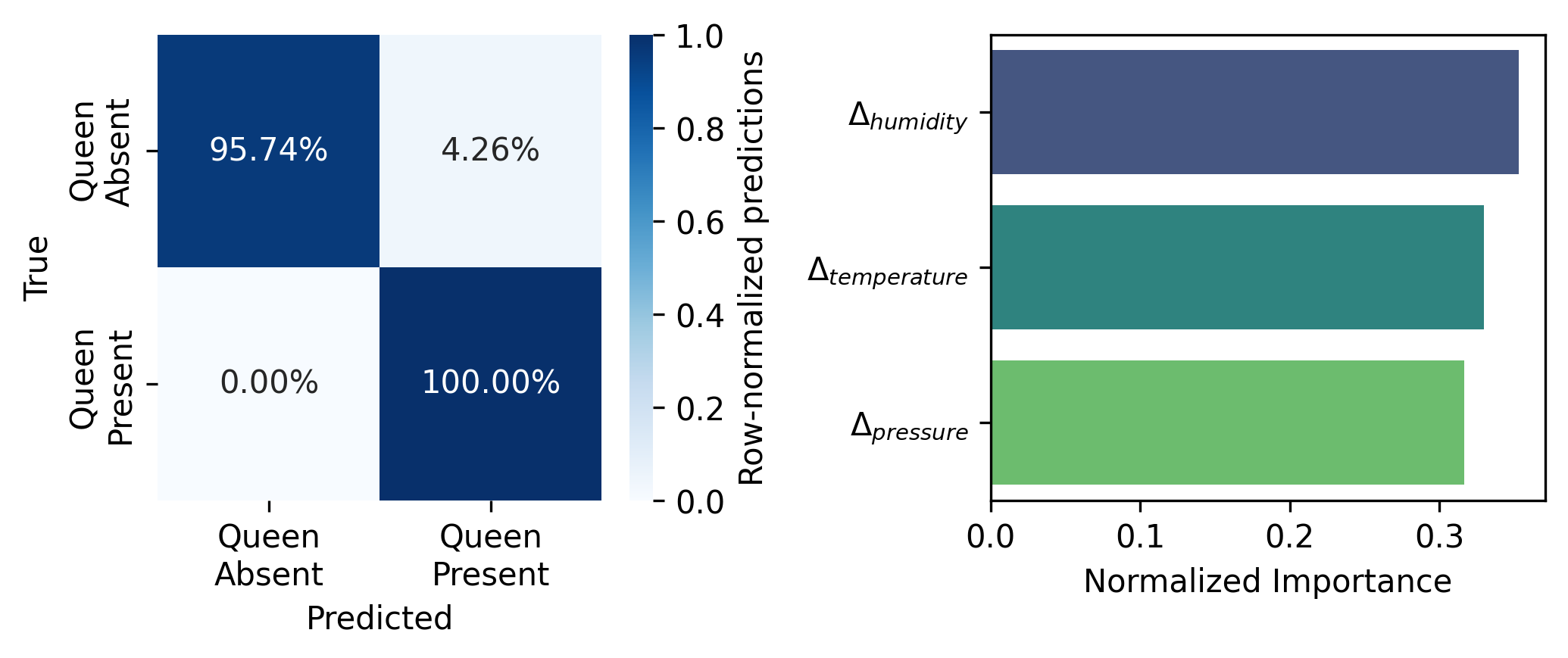}
    \caption{Classification performance of the trained model in full resolution using only environmental information as input features. (Left) Row-normalized confusion matrix showing classification results between the "Queen Present" and "Queen Absent" classes. The model achieves perfect recall for detecting the queen's presence and a high true negative rate for her absence. (Right) Feature importance plot showing the normalized contribution of sensor-derived features (\(\Delta\)humidity, \(\Delta\)temperature, \(\Delta\)pressure) to the classification decision. Humidity variation appears to be the most influential feature.}
    \label{fig:accuracy-duration}
\end{figure}
\subsection{Minimal impact of audio features}
\label{ssec:audio}
In addition to environmental features, we evaluated models that included a basic audio feature: the Root Mean Square (RMS) energy computed over the 200--700~Hz frequency band on the 60s of acquisition. This feature was selected for its computational simplicity and suitability for online processing, as it does not require buffering or storing large volumes of audio data. Its lightweight nature makes it compatible with low-power embedded systems.
When combined with environmental data, the inclusion of RMS energy resulted in only a marginal improvement in test accuracy (up to 98.7\%), falling short of the performance achieved using environmental features alone (99.4\%). 
Given the minimal benefit and higher resource demands, we conclude that audio sensing is not justified in this context, especially for embedded, energy-constrained applications.

\subsection{STM32 performance evaluation}
\label{ssec:stm32}
The final evaluation was conducted on a total of $1420$ test files, with both Python and STM32 implementations demonstrating good classification performance depicted in Table~\ref{tab:stm-performance}. The former achieved $1410$ correct predictions, resulting in an accuracy of 99.3\%, while the STM32 implementation closely followed with an accuracy of 99.2\% (1,408 correct predictions). In terms of power consumption, using a limited set of environmental features has only a minor effect on the size of the network and the overall energy demand, since the number of input variables to process remains relatively small. In contrast, incorporating audio data substantially increases the volume of streaming input, leading to a significant rise in power consumption (almost a twofold increase) compared to the environmental-only configurations.
%
%
Resource-wise, the STM32 implementation maintains a compact memory footprint, with estimated sensor data usage around 24 bytes and model size approximately 10,000 bytes, totaling roughly 10,024 bytes of RAM usage. CPU utilization was relatively high at 92.7\%, indicative of intensive computation but acceptable within embedded constraints.
Confusion matrices show both implementations perform very well, with zero false negatives and only a slight difference in false positives. Classification reports confirm these results, with macro-average F1 scores near 0.98 and weighted averages close to 0.99. The higher false positive count on STM32 likely stems from limited numerical precision and differences in floating-point handling or feature scaling; further tuning—such as refining quantization or improving preprocessing—could reduce this gap and align performance more closely with the Python baseline.
Overall, these results demonstrate that the STM32 implementation closely matches the model’s performance while operating efficiently within embedded system constraints, making it a viable option for real-time classification tasks in resource-limited environments.

\subsection{Comparison with other works}
\label{ssec:comparison}
Table~\ref{tab:comparison} compares recent queen detection systems in terms of sensing, accuracy, platform, and energy efficiency. Audio-based approaches like De Simone et al. and Barbisan et al. achieve high accuracy (96–98\%) but depend on complex preprocessing and consume up to 75 mJ per inference. Multi-sensor platforms offer valuable datasets and architecture insights, but lack real-time embedded implementation or power metrics~\cite{Zhu_eta24, borowik2019multisensor, Cota_etal23}.
In contrast, our system achieves $>$99\% accuracy using only environmental sensors, with an energy cost of just $\sim$6.45 mJ per inference on an STM32 NUCLEO-F767ZI. This marks a $>$11X energy improvement over the most efficient prior method, while enabling fully embedded, low-power, and scalable deployment for autonomous hive monitoring.

\section{Conclusions}
This work presents a lightweight and energy-efficient system for real-time queen bee presence detection using environmental sensor fusion and embedded machine learning.
%
The STM32 implementation maintained nearly identical performance to the Python reference model, with sub-second inference latency and an energy cost under 10~mJ per prediction—making it suitable for long-term deployment in constrained environments. Compared to existing audio-based approaches, our system offers a more scalable and sustainable solution, enabling non-invasive hive monitoring for precision beekeeping.
While our method is non-invasive in terms of colony disturbance, it does require the integration of environmental sensors inside the hive; such sensors can be unobtrusively embedded in standard beekeeping setups, and are generally robust to hive conditions such as high humidity and moderate wax accumulation, though long-term deployments may require periodic maintenance to ensure measurement reliability under challenging environmental and weather conditions.
Future work may explore adaptive sampling, extended sensor modalities, and swarm-level integration, further advancing autonomous ecological sensing in practice.

\section*{Acknowledgment}
C.D.L. acknowledge the financial support of the Bridge Fellowship founded by the Digital Society Initiative at University of Zurich (Grant No. G-95017-01-12). 

\printbibliography

\end{document}